# A Voxel-based Approach for Simulating Microbial Decomposition in Soil: Comparison with LBM and Improvement of Morphological Models

Mouad KLAI[1,2], Olivier Monga[1,2], Mohamed Soufiane JOUINI [3], and Valérie POT [4]

[1] Laboratory of Mathematics and Population Dynamics (LMDP), Cadi Ayyad University, Marrakech, Morocco.
[2] Unit for Mathematical and Computer Modeling of Complex Systems (UMMISCO), IRD, Sorbonne University, Paris, France.
[3] Mathematics Department, Khalifa University of Science and Technology, Abu Dhabi, United Arab Emirates.
[4] Université Paris-Saclay, INRAE, AgroParisTech, UMR ECOSYS, Palaiseau, France.

Corresponding author: Soufiane JOUINI (e-mail : mohamed.jouini@ku.ac.ae)

**ABSTRACT** This study presents a new computational approach for simulating the microbial decomposition of organic matter, from 3D micro-computed tomography (micro-CT) images of soil. The method employs a valuated graph of connected voxels to simulate transformation and diffusion processes involved in microbial decomposition within the complex soil matrix. The resulting model can be adapted to simulate any diffusion-transformation processes in porous media. We implemented parallelization strategies and explored different numerical methods, including implicit, explicit, synchronous, and asynchronous schemes. To validate our method, we compared simulation outputs with those provided by LBioS and by Mosaic models. LBioS uses a lattice-Boltzmann method for diffusion and Mosaic takes benefit of Pore Network Geometrical Modelling (PNGM) by means of geometrical primitives such as spheres and ellipsoids. This approach achieved comparable results to traditional LBM-based simulations, but required only one-fourth of the computing time. Compared to Mosaic simulation, the proposed method is slower but more accurate and does not require any calibration. Furthermore, we present a theoretical framework and an application example to enhance PNGM-based simulations. This is accomplished by approximating the diffusional conductance coefficients using stochastic gradient descent and data generated by the current approach.

**INDEX TERMS** Microbial Decomposition Simulation, Flow Simulation, 3D Micro-Computed-Tomography Images, Lattice Boltzmann Method, Pore Space Geometrical Modelling, Diffusional Conductance Coefficients,

**CREDIT AUTHOR STATEMENT**
**Mouad KLAI:** Conceptualization, Methodology, Implementation, Formal analysis, Visualization, Writing – Original Draft.
**Olivier MONGA:** Supervision, Conceptualization, Methodology, Implementation, PNGMS, Writing – Review & Editing.
**Valerie POT:** LBM implementation, Writing – Review & Editing.
**Mohamed Soufiane Jouini:** Writing – Review & Editing.

## I. INTRODUCTION

Microbial activity in soil is essential for maintaining soil health and supporting ecosystem functioning. Microorganisms contribute to nutrient cycling, organic matter decomposition, and plant growth promotion [1],[2]. Soil demonstrates high diversity and complex dynamic interactions, making it challenging to fully capture its complexity. Moreover, as a heterogeneous and intricate medium, it poses difficulties for directly observing and measuring microbial activity [3]. Traditional approaches, such as culturing and microscopy are time-consuming and provide limited insights. Therefore, innovative and advanced numerical simulation techniques are needed to study soil microbial activity [4] - [24].

Computed Tomography (CT) imagery offers a non-invasive method for visualizing porous media structures at the microscale of the soil microbial habitats. Currently, standard 3D CT imaging generates large image data volumes, often starting from around one billion voxels [10] - [12].

In the last 20 years, many studies have focused on improving computational models for simulating dynamics from 3D CT images of soil. These efforts aim to understand how soil components move and interact in porous media and in fractured environments. Several studies have reported significant contributions, where models typically integrate a transport mechanism with advanced reaction (transformation) processes [13] - [16].

Different methodologies, such as: Lattice-Boltzmann Method (LBM), smooth particle hydrodynamics [14], hybrid Lattice Boltzmann-direct numerical simulation (DNS) [15], and pore network geometrical models (also known as morphological models) [16], are employed to describe chemical transport phenomena.

LBM based approaches have been widely adopted to simulate microbial activity in porous media using micro 3D CT soil images [17] - [19]. However, LBM is very computationally expensive and require large memory capabilities. For instance, with a regular laptop, a five-day (real time) simulation of microbial decomposition of organic matter with an image size of $512^3$ voxels with 17% porosity takes about 3 weeks [20].

Alternatively, Pore Network Geometrical Modeling (PNGM)-based simulation offers a cost-effective tool with reduced computing time and moderates memory requirements. These models represent pore space by a valuated graph of connected primitives such as balls, ellipsoids, or cylinders [20] - [23].

In PNGM-based simulations, calibration is essential to fine-tune the transport simulation, especially by adjusting diffusional conductance coefficients for connected pairs of primitives [24] - [26].

This paper introduces a computational method for simulating microbial activity in soil from 3D CT images of pore space, departing from previous approaches that rely on modeling pore space using geometrical primitives. Our method improves simulation accuracy by representing the pore space as a graph of connected voxels, and modeling transport phenomena using Fick's law of diffusion [27]; resulting in a more adaptable simulation tool that does not require calibration. We also implement parallelization strategies and explore different numerical schemes, including implicit, explicit, synchronous, and asynchronous schemes.

Furthermore, we introduce a theoretical framework to improve diffusion simulation in pore network geometrical modeling by approximating diffusional conductance coefficients for connected pairs of primitives. In addition, we provide an application example for improving simulation of diffusion and microbial decomposition of organic matter in soil using a ball network model.

Throughout this paper, we provide a theoretical framework and a comparative analysis of simulation results on a 3D micro-CT image of real sandy loam soil samples.

The remainder of this paper is structured as follows. In section II, we illustrate our approach, including mathematical formulation, and numerical schemes. The implementation details are discussed within Section III. In section IV, we present a comparative analysis of the present approach with Mosaic and LBM-based simulations. In section V, we use the present approach to improve PNGM-based simulation by approximating diffusional conductance coefficients of connected primitives. Finally, we discuss the advantages and perspectives of our method in section VI.

## II. MODELING STRATEGY AND SIMULATION FRAMEWORK

### A. NOTATIONS AND PRINCIPLE OF SIMULATION

In this study, we assume that we are working in an aquatic environment (100% saturation), and that the pore space is filled with water. We model the dynamics of microbial decomposition using five key compounds, as illustrated in Figure 1:

- Microbial biomass (MB): Represents the mass of microorganisms in the sample.

- Microbial respiration ($CO_2$): Indicates the produced carbon dioxide through microbial decomposition, indicative of microbial growth.

- Fresh organic matter (FOM): Derived from recently added or deposited plant and easily decomposable

- Soil organic matter (SOM): Consists of various organic compounds in different stages of decomposition, originating from biomass turnover and less accessible to decomposition.

- Dissolved organic matter (DOM): Refers to organic compounds dissolved in soil water, originating from hydrolyze of FOM and SOM and biomass recycling, available for microbial uptake or transport within the pore space.

Note that for environments where saturation is not complete, the same simulation framework can be used after a method of drainage [17], [18].

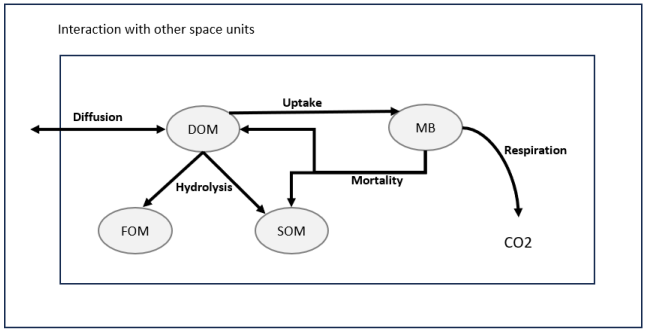

FIGURE 1. The processes involved within the microbial decomposition of organic matter in soil: DOM comes from the decomposition of SOM (slow decomposition) and FOM (fast decomposition). The microorganisms grow by assimilating DOM, breathe by producing CO2 and when they die they are recycled into DOM and SOM.

Let $(I(i,j,k))_{i,j,k}$ be a 3D binary image where the voxels forming pore space (void voxels) are tagged by 0 and the voxels attached to solid matter (solid voxels) are tagged by 1. Let $V = \{(i,j,k) : I(i,j,k) = 0\} = \{v_1, ..., v_n\}$ be the set of pore space voxels and $N = \{1, ..., n\}$ the index set of $V$, where $n = card(V)$ represents the number of voxels of the pore space.

We construct an adjacency valuated graph $G(V, E)$ where $V$ is the set of nodes, and $E = \{(i,j) \in N^2 | v_i \cap v_j \neq \emptyset\}$ is the set of edges. In this work, we use 6-connecvity between voxels (Figure 2)

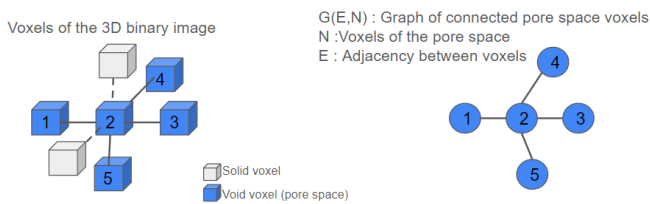

FIGURE 2. The Graph construction using 6-connectivity policy.

Let $t$ be the variable representing time. To each voxel $v_i$ of the pore space, we attach: the mass of MB ($x_i^1(t)$), the mass of DOM ($x_i^2(t)$), the mass of SOM ($x_i^3(t)$), the mass of FOM ($x_i^4(t)$), the mass of CO2 ($x_i^5(t)$).

To each node $v_i$ let us denote by $X_i(t) = (x_i^1(t), x_i^2(t), x_i^3(t), x_i^4(t), x_i^5(t))$ the masses of the compounds contained within voxel $v_i$ at time t.

We simulate microbial activity by updating the constructed graph according to transformation and diffusion processes described in Figure 3. The principle is to break down the complex processes, that can be mathematically modeled, into simpler steps.

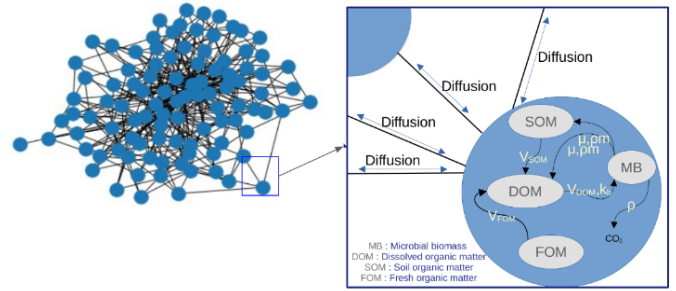

FIGURE 3. The Graph representation of microbial activity model.

Let us assume that we get a representation of the pore space and its compounds at time $t$ in the form of $G_t(V, E, (X_i(t))_{i \in N})$.

To get $G_{t+\delta t}(V, E, (X_i(t + \delta t))_{i \in N})$ at time $t + \delta t$, after microbial activity processes we apply a transformation model that encodes the biological conversion of different compounds within each voxel separately. Afterward, we apply a discretized model to mimic the diffusion of compounds between the connected voxels.

B. Transport modeling and graph diffusion equation
1) Fick's law of diffusion

Fick's law of diffusion describes the process of molecular or particle movement through a medium, such as gases, liquids, or solids. The law, formulated by Adolf Fick, states that the rate of diffusion of a substance is directly proportional to the concentration gradient of that substance [27]. In simpler terms, it asserts that substances naturally move from areas of higher concentration to areas of lower concentration. This fundamental principle is mathematically expressed as:

$$J = -D \frac{dC}{dx}$$

Where:
$J$ represents the diffusion flux (amount of substance per unit area per unit time), $D$ is the diffusion coefficient of the substance and $dc/dx$ the concentration gradient along the diffusion direction.

This law is widely applicable and plays a crucial role in understanding various natural processes, such as the diffusion of gases in biological systems, the movement of ions in electrolytes, and the transport of solutes in porous materials.

2) Diffusion Modeling: Graph Diffusion Equation
Within this work, we use Fick's Law of Diffusion to model the mass exchanges between voxels by updating the features (masses compounds) of the graph nodes.

Let's consider diffusion process of DOM. Let $m_i(t) = x_i^2(t)$ represent the mass of DOM at time $t$ in the voxel $p_i$. Considering the voxel as the unit of our framework, where the volume of a voxel is equal to one, the concentration at a voxel equals its mass.

We denote $\delta m_{i,j}$ as the difference in concentration between the voxel $p_i$ and its neighboring voxel $p_j$:
$$\delta c_{i,j}(t) = \delta m_{i,j}(t) = m_i(t) - m_j(t)$$
The mass flow between time $t$ and time $t + \delta t$ from the voxel at position $p_i$ to its neighboring voxel at position $p_j$ is given by the first Fick's law as follows:
$$J_{i,j} = -\frac{D_{DOM}.S_{i,j}.\delta c_{i,j}(t)}{d_{i,j}}.\delta t$$
where $D_{DOM}$ is the DOM diffusion coefficient in water, $S_{i,j}$ is the contact surface between the two voxels, and $d_{i,j}$ is the Euclidian distance between the two centers of inertia of the voxels. Since the unit of the framework is the voxel, $S_{i,j}$ and $d_{i,j}$ are reduced to one. Since $\delta c_{i,j}(t) = \delta m_{i,j}(t)$, the mass flow between the two voxels during a period $\delta t$ is:
$$J_{i,j} = -D_{DOM} . \delta m_{i,j}(t). \delta t$$
Thus, the variation of DOM mass between time $t$ and time $t + \delta t$ due to diffusion is,
$$m_i(t + \delta t) - m_i(t) = \sum_{j:(i,j)\in E} -D_{DOM}\, \delta m_{i,j}(t)\, \delta t$$
.
Then the governing equation is:
$$\frac{dm_i(t)}{dt} = \sum_{j:(i,j)\in E} -D_{DOM}\left(m_i(t) - m_j(t)\right)$$
Gathering all the masses of all the voxels of the pore space in one vector $M(t) = (m_1(t) \dots m_n(t))$ we get
$$\frac{dM(t)}{dt} = -D_{DOM}.\Delta M(t) \quad (1)$$
Where $\Delta = (\delta_{i,j})_{1\leq i,j\leq n}$ is the Laplacian matrix of the graph $G(E,N)$ defined by:
$$\begin{cases} \delta_{i,j} = deg(i) & if\ i = j \\ \delta_{i,j} = -1 & if\ i \neq j\ and\ (i,j) \in E \\ \delta_{i,j} = 0 & otherwise \end{cases}$$
where $deg(i) = \sum_{j:(i,j)\in E} 1$ is the number of adjacent nodes to the node i.
Equivalently, $\Delta = D - A$ where $D$ is the degree matrix and $A$ is the graph's adjacency matrix of the graph $G(N,E)$. The equation 1 is called Graph Diffusion Equation (GDE) of $G(N,E)$.
The graph diffusion equation is linear, and since the eigenvalues of the graph Laplacian $\Delta$ are positive because it is a symmetric and diagonally dominant matrix, and hence the eigenvalues of the matrix $-D_{DOM}\Delta$ are non-positive. Consequently, the ODE is stable. In addition, it has the solution
$$M(t) = e^{-D_{DOM}\Delta\, t}.M(0)$$
Performing eigenvalue decomposition, the solution is
$$M(t) = \Gamma. e^{-D_{DOM}.t.\Psi}.\Gamma^{-1}.M(0)$$
Here $\Gamma$ and $\Gamma^{-1}$ are matrices related to the eigenvectors of the system, and $\Psi$ is a diagonal matrix containing the eigenvalues.
Assuming $\Gamma^{-1}$ exists, $\Gamma$ has full rank and both are bounded, the equation becomes
$$y(t) = e^{-D_{DOM}\Psi\, t}y(0)$$

where $y(t) = \Gamma M(t)$.
If $M(t)$ and $\widehat{M}(t)$ are two solutions of the ODE then their projections in the eigenspace are $y(t)$ and $\hat{y}(t)$. For each node $i \in N$ we have:
$$|y_i(t) - \hat{y}_i(t)| = |(y_i(0) - \hat{y}_i(0))e^{\underline{\lambda}_i t}|$$
$$= |(y_i(0) - \hat{y}_i(0))|e^{Re(\underline{\lambda}_i)t}$$
for this to converge to zero as $t \to \infty$ we require $Re(\underline{\lambda}_i) \leq 0 \ \forall\ i \in N$, which is the case.

3) *Numerical schemes*
Solving the graph diffusion equation (1) consists of solving the following set of equations (2) for all nodes of the graph simultaneously,
$$\forall i \in N,\ \frac{dm_i(t)}{dt} = \sum_{j:(i,j)\in E} -D_{DOM}\,(m_i(t) - m_j(t)) \quad (2)$$

   a) *Explicit scheme*

Let $i \in N$, The simplest way to discretize the first derivative with respect to time is by using the forward Euler time scheme, namely:
$$\frac{dm_i(t)}{dt} = \frac{m_i(t + \delta t) - m_i(t)}{\delta t}$$
Then we have:
$$\frac{m_i^{(k+1)} - m_i^{(k)}}{\delta t} = \sum_{j:(i,j)\in E} D_{DOM}\,(m_i^{(k)} - m_j^{(k)})$$
where $k$ is the iteration number (i.e., the discrete-time index), and $\delta t$ is the time step.
Equivalently, we have:
$$m_i^{(k+1)} = m_i^{(k)} - \sum_{j:(i,j)\in E} D_{DOM}\delta t\,(m_i^{(k)} - m_j^{(k)})$$
Rewritten compactly in matrix-vector form as follows,
$$M^{(k+1)} = (I - D_{DOM}\delta t\, \Delta)M^{(k)}$$
$$= Q\, M^{(k)} \quad (\text{Scheme 1})$$
where the matrix $Q = (q_{i,j})_{1\leq i,j\leq n}$ is given by;
$$\begin{cases} q_{i,j} = 1 - D_{DOM}\delta t\, deg(i) & if\ i = j \\ q_{i,j} = D_{DOM}\delta t & if\ i \neq j\ and\ (i,j) \in E \\ q_{i,j} = 0 & otherwise \end{cases}$$
This approach is termed explicit because the update $M^{(k+1)}$ is directly derived from $M^{(k)}$ through the application of the diffusion operator $Q$. The solution to the diffusion equation is obtained by iteratively applying the scheme 1 multiple times in succession, starting from an initial distribution $M^{(0)}$.

   b) *Implicit Scheme*

Let $i \in N$, another way to discretize the first derivative with respect to time is by using a backward Euler time scheme, namely:
$$\frac{m_i^{(k+1)} - m_i^{(k)}}{\delta t}$$
$$= \sum_{j:(i,j)\in E} -D_{DOM}\,(m_i^{(k+1)} - m_j^{(k+1)})$$
Then we have:

$$m_i^{(k)} = m_i^{(k+1)} + \sum_{j:(i,j)\in E} D_{DOM}\, \delta t\, (m_i^{(k+1)} - m_j^{(k+1)})$$

Rewritten compactly in matrix-vector form,
$$M^{(k)} = (I + D_{DOM}\delta t\, \Delta)\, M^{(k+1)} \quad \text{(Scheme 2)}$$
$$= B\, M^{(k+1)}$$

where the matrix $B = (b_{i,j})_{1\leq i,j \leq n}$ is given by:
$$\begin{cases} b_{i,j} = 1 + D_{DOM}\delta t\, deg(i) & \text{if } i = j \\ b_{i,j} = -D_{DOM}\delta t & \text{if } i \neq j \text{ and } (i,j) \in E \\ b_{i,j} = 0 & \text{otherwise} \end{cases}$$

This scheme is referred to as implicit because it involves resolving a linear system to calculate the state $M^{(k+1)}$ based on $M^{(k)}$, which necessitates the inversion of the matrix $B$.

In practice, rather than exact inversion, a few iterations of a linear solver are commonly employed in place of inverting the matrix.

Given that $B$ is a large, sparse, and symmetric positive definite matrix, the Conjugate Gradient method stands out as the optimal iterative solver for addressing the scheme 2. The Conjugate Gradient method is particularly effective for symmetric positive definite matrices, as it leverages conjugate directions to minimize the error between the candidate solution and the target iteratively. In [20], to solve an implicit scheme for simulating diffusion using Fick's first law of diffusion in a graph of connected spheres approximating the pore space, the conjugate gradient method was employed. This method has shown advantages in scenarios involving large-scale, sparse linear systems, with its efficiency further enhanced when a suitable preconditioner is employed.

### C. Modeling microbial decomposition of organic matter processes in the graph of connected voxels

Microbial activity is governed not only by the diffusion of various compounds within the soil's pores but also by transformation processes. Dissolved Organic Matter (DOM) arises from the decomposition of both slow-decomposing Soil Organic Matter (SOM) and fast-decomposing Fresh Organic Matter (FOM). Microorganisms grow through the assimilation of DOM and respire by producing $CO_2$. Upon death, they are transformed back into organic matter.

We recall that for each voxel at position $p_i$:
- $x_i^1(t)$ is the mass of MB,
- $x_i^2(t)$ is the mass of DOM,
- $x_i^3(t)$ is the mass of SOM,
- $x_i^4(t)$ is the mass of FOM,
- $x_i^5(t)$ is the mass of $CO_2$.

Let $X^j(t) = \left(x_1^j(t), \cdots, x_n^j(t)\right)$ represent the distribution of the $j$th biological parameter in the pore space.
For instance, $X^1(t) = \left(x_1^1(t), \cdots, x_n^1(t)\right)$ is the mass distribution of microorganisms in the pore space.
Let $v_i \in V$ be a node of the graph.

We model the growth of microorganisms by consuming the available dissolved organic matter according to the Monod equation. Exponential models are employed to represent both death and mass loss due to respiration through mineralized organic matter emission. The total variation of $x_i^1(t)$ is described by the following equation:
$$\frac{dx_i^1(t)}{dt} = -\rho . x_i^1(t) - \mu . x_i^1(t) + \frac{v_{DOM} . x_i^2(t)}{K_{DOM} + x_i^2(t)} . x_i^1(t)$$

Where $\rho$ is the respiration rate, $\mu$ is the mortality rate, $v_{DOM}$ and $K_{DOM}$ are respectively the maximum growth rate of MB and the constant of half saturation of DOM.

In the saturated pore space, dissolved organic matter comes from dead microorganisms, the transformation of soil organic matter, and fresh organic matter. As before, a portion is allocated from DOM to microorganisms according to the Monod equation. Furthermore, in addition to diffusion processes, the total variation of $x_i^2(t)$ is described by the following equation:

$$\frac{dx_i^2(t)}{dt} = -D_{DOM}.[\Delta X^2(t)]_i + \beta.\mu.x_i^1(t)$$
$$- \frac{v_{DOM}.x_i^2(t)}{K_{DOM} + x_i^2(t)}.x_i^1(t) + v_{SOM}.x_i^3(t)$$
$$+ v_{FOM}.x_i^4(t)$$

Where $\Delta = (\delta_{i,j})_{1\leq i,j \leq n}$ is the Laplacian matrix of the graph $G(V,E)$, $[\Delta X^2(t)]_i$ is the $i$th component of $\Delta X^2(t)$, $D_{DOM}$ is the diffusion coefficient of DOM in water, $\beta$ is the proportion of MB returning to DOM (the other fraction $(1 - \beta)$ returns to SOM), $v_{FOM}$ and $v_{SOM}$ the hydrolysis rate of FOM and SOM.

A part of soil organic matter comes from dead microorganisms so the total variation of $x_i^3(t)$ is described by the following equation:
$$\frac{dx_i^3(t)}{dt} = (1 - \beta)\mu.x_i^1(t) - v_{SOM}.x_i^3(t)$$

The density evolution of FOM ($x_i^4(t)$) is presented by:
$$\frac{dx_i^4(t)}{dt} = -v_{FOM}.x_i^4(t)$$

The evolution of $CO_2$ ($x_i^5(t)$) is governed by microorganisms breathing, which leads to the following equations:
$$\frac{dx_i^5(t)}{dt} = \rho.x_i^1(t).$$

The present models for microbial activity in soil have been validated in [28].

By gathering all the equations, we have the following system of equations:

$$\begin{cases}
\frac{dx_i^1(t)}{dt} = -\rho \cdot x_i^1(t) - \mu \cdot x_i^1(t) \\
\qquad + \frac{v_{DOM} \cdot x_i^2(t)}{K_{DOM} + x_i^2(t)} \cdot x_i^1(t) \\
\frac{dx_i^2(t)}{dt} = -D_{DOM} \cdot [\Delta X^2(t)]_i + \beta \cdot \mu \cdot x_i^1(t) - \frac{v_{DOM} \cdot x_i^2(t)}{K_{DOM} + x_i^2(t)} \cdot x_i^1(t) \\
\qquad + v_{SOM} \cdot x_i^3(t) + v_{FOM} \cdot x_i^4(t) \\
\frac{dx_i^3(t)}{dt} = (1-\beta)\mu \cdot x_i^1(t) - v_{SOM} \cdot x_i^3(t) \\
\frac{dx_i^4(t)}{dt} = -v_{FOM} \cdot x_i^4(t) \\
\frac{dx_i^5(t)}{dt} = \rho \cdot x_i^1(t)
\end{cases}$$

By removing the diffusion processes of DOM from the equations, and applying an explicit scheme to the first-order derivatives of the equations we obtain the following equations for the transformation processes:

$$\begin{cases}
x_i^1(t+\delta t) = x_i^1(t) + \frac{v_{DOM} \cdot x_i^2(t)}{K_{DOM} + x_i^2(t)} \cdot x_i^1(t) \cdot \delta t \\
\qquad - (\rho \cdot x_i^1(t) + \mu \cdot x_i^1(t)) \cdot \delta t \\
x_i^2(t+\delta t) = x_i^2(t) - \frac{v_{DOM} \cdot x_i^2(t)}{K_{DOM} + x_i^2(t)} \cdot x_i^1(t) \cdot \delta t \\
\qquad + (v_{SOM} \cdot x_i^3(t) + v_{FOM} \cdot x_i^4(t) + \beta \cdot \mu \cdot x_i^1(t)) \cdot \delta t \\
x_i^3(t+\delta t) = x_i^3(t) + ((1-\beta)\mu \cdot x_i^1(t) - v_{SOM} \cdot x_i^3(t)) \cdot \delta t \\
x_i^4(t+\delta t) = x_i^4(t) - v_{FOM} \cdot x_i^4(t) \cdot \delta t \\
x_i^5(t+\delta t) = x_i^5(t) + \rho \cdot x_i^1(t) \cdot \delta t
\end{cases} \quad (3)$$

To obtain the voxel states after a time step $\delta t$ of transformation processes, we can apply the equations synchronously or asynchronously. Given that the transformation of the nodes is independent, we have the flexibility to perform transformations sequentially or in parallel on the graph's nodes.

In the next section, we provide a comprehensive explanation of the implementation of all discussed concepts. This includes details on graph construction from the 3D image, the conjugate gradient method, its parallelization, as well as the implementation of both implicit and explicit schemes. Furthermore, we provide details of implementing each transformation scheme, whether synchronous or asynchronous.

## III. IMPLEMENTATION DETAILS

Generally, simulating transformation-diffusion processes in the complex geometry captured by the 3D image involves transforming the pore space in the image into a graph of connected voxels. Subsequently, each simulation iteration applies transformation processes to the biological attributes of all graph nodes, followed by the solution of the graph diffusion equation to mimic diffusion of the compound to be diffused in the model (DOM in this study). Transforming the 3D image into a graph of connected voxels is a straightforward and parallelizable process. It can be accomplished by enumerating the valid voxels in the image (i.e., those corresponding to the pore space). Then, looping over the voxels of the image, when we encounter a valid voxel, we identify from the 6 neighboring voxels (Figure 2) the valid ones and construct the adjacency graph.

Let $t \geq 0$ be a specific time, and let $(X_i(t))_{i \in N}$ denote the distribution of the nodes at time $t$. The attributes of the nodes, i.e., the distribution of the pore space $(X_i(t+\delta t))_{i \in N}$ after a time step $\delta t$ of microbial activity, are obtained by applying the processes in system S1 synchronously (Algorithm 1 of annex A) or asynchronously (Algorithm 2 of annex A). Then, the explicit scheme (Scheme 1) or implicit scheme (Scheme 2) is applied in order to obtain the distribution after diffusion processes. In the synchronous procedure for performing microbial transformation, we calculate the contribution of biological variables to uptake, respiration, mortality, and turnover for each node. Subsequently, we update the biological variables according to the logic of system 3. In contrast, the asynchronous procedure involves sequentially updating the biological variables of the nodes. This means allowing microorganisms to grow by consuming DOM, followed by a portion dying then respiring to produce CO2, and transforming dead microbial biomass to DOM and SOM. Detailed algorithms for both synchronous and asynchronous transformations are described in Appendix A.

When the time steps for both transformation and diffusion are identical, the processes are performed just once per time step. However, if the time steps differ between these processes, we iterate through the process (transformation or diffusion) with the smallest time step to ensure that the dynamics of the entire transformation-diffusion system remain accurate.

When using the explicit scheme, the computational process involves straightforward matrix multiplication, which is relatively efficient. On the other hand, implementing the implicit scheme requires solving a linear system that involves a large, sparse, symmetric, and positive definite matrix. To handle such computations effectively, we employ the preconditioned conjugate gradient method, where the conditioning matrix $= (t_{i,j})_{0 \leq i,j \leq n}$, is defined by

$$\begin{cases} t_{i,j} = \frac{1}{b_{i,j}} & \text{if } i = j, \\ t_{i,j} = 0 & \text{else} \end{cases}$$

Where $B = (b_{i,j})_{0 \leq i,j \leq n}$ is the matrix of the implicit scheme defined in subsection 1.2.3.2.

The Preconditioned Conjugate Gradient (PCG) method is a variant of the Conjugate Gradient (CG) method used for solving linear systems of equations $Ax = b$, where $A$ is a symmetric positive definite (SPD) matrix. The PCG method incorporates a preconditioner matrix $T$ to improve convergence speed.

The implementation was done using the C language due to its capacity for handling such complex problems and its efficient memory management and its computational speed required to address some aspects of the simulation like: graph construction from the 3D image, solving the graph diffusion equation through the numerical schemes, running expensive simulations on the constructed graph.

## IV. VALIDATION USING LBM-BASED SIMULATION AND COMPARISON WITH PNGM-BASED SIMULATION

In this section, we compare simulations using the presented approach that we call Voxel Graph-based Approach (VGA), with LBM-based simulation and pore network-based simulations. For LBM-based simulation, we use the approach outlined in [22], where diffusion is computed from the lattice-Boltzmann equation and transformation is calculated using the synchronous scheme.

For the pore network model, we use the generic tool discussed in [20], where we model the pore space using a minimal set of maximal balls covering the complex shape of the pore space, then a graph of connected balls is constructed, and diffusion is modeled using the first Fick's law that we calibrated using the LBM approach.

### A. Dataset description

For comparison purposes, we use the data taken from [4] and used in [7] and [20]. The resolution in the image, with a pixel size of 24 $\mu m$, is sufficiently high to depict various pore classes associated with microbiological activity. Additionally, it is sufficiently small to model diffusion within a millimetric size of soil volume.

The soil used in this investigation originates from the Bullion Field, an experimental site at the James Hutton Institute in Invergowrie, Scotland. It is characterized as sandy loam soil with 71% sand, 19% silt, and 10% clay by soil mass [4].

For the computational simulations, a volume image, of $512^3$ voxels, was extracted from the 3D stack, equivalent to a soil sample volume of approximately $1.855\ cm^3$. The separation of solid and pore phases was achieved by applying an indicator kriging method [29].

The resultant segmented 3D binary image indicates a visible porosity of 17% for the soil sample with a bulk density of 1.2 g cm$^{-3}$ and 8% for the soil sample with a bulk density of 1.6 g cm$^{-3}$.

Figure 4 shows random cross sections of the 3D binary image. Figure 5 shows a 3D view of the binary image using MATLAB routine.

To perform simulations based on pore network models, we extract from the 3D binary image the minimal set of maximal spheres approximating the pore space [21]. Figure 6 depicts a view of the ball network wherein we specifically choose the balls whose centers reside within the region bounded by $[[50, 100], [50, 100], [150, 200]]$.

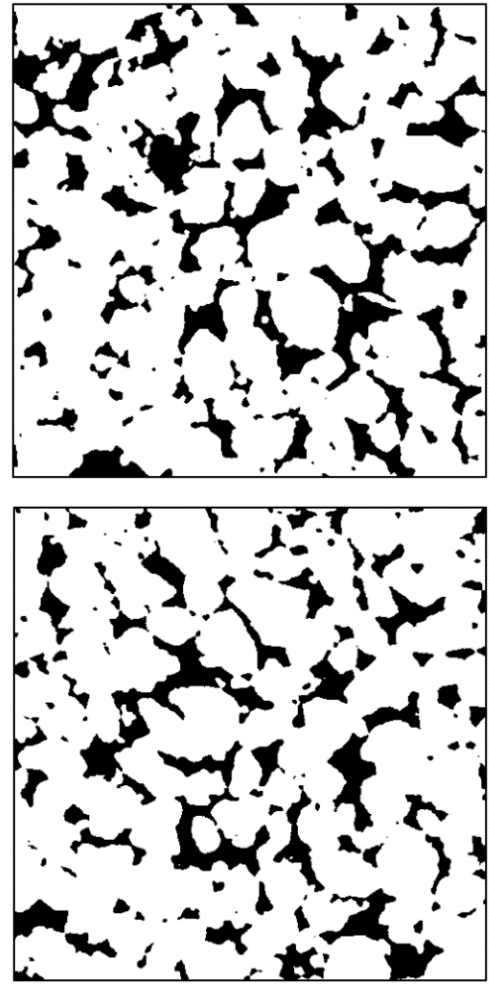

FIGURE 4. Random cross sections of the 3D binary image.

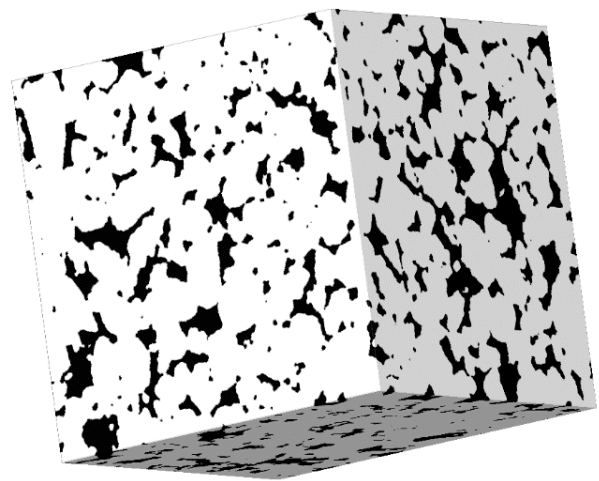

FIGURE 5. 3D view of the 3D binary image.

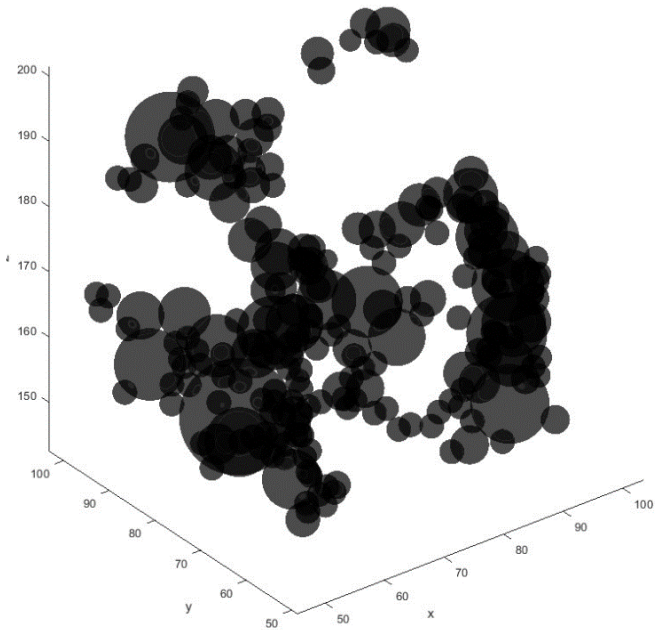

**FIGURE 6.** Balls where centers reside within the region bounded by $[[50, 100], [50, 100], [150, 200]]]$.

### B. Comparison of diffusion simulations: VGA, LBM and PNGM-based simulations

To compare diffusion simulation using the graph diffusion equation with simulations using the Lattice Boltzmann method and the pore network model, we draw upon a prior experience detailed in [20], which was conducted to compare diffusion using a pore network model (ball network model) and LBM, and we outline our approach as follows:

Uniformly, we distributed $M_0 = 592.7593\ mg$ of DOM on the first two z layers ($z = 1$ and $z = 2$) of the $512 \times 512 \times 512$ image. In our approach and the LBM, we selected all the voxels of the two layers ($z = 0$ and $z = 1$) corresponding to the pore space and distribute $M_0$ uniformly among the selected voxels. For the ball network model, we selected the balls intersecting the two layers and distributed the mass $M_0$ homogeneously across them (i.e., applying the same concentration to all the balls).

The experiment aims to compare mass profiles in each layer of the image after a time period of $1.76\ hours$. We run the simulations using the three methods and obtained the final distributions, from which we calculated the mass in each layer of the image. Subsequently, we plot these results together for comparison.

The basic principle of pore network geometrical modeling (PNGM) is to globally transport mass from one ball to the connected balls in accordance with Fick's laws. Therefore, we need to calibrate the diffusional conductance [29]. We employed a general coefficient $\alpha = 0.6$, that we multiplied by the contact surface area between each connected ball, calibrated from LBM [20].

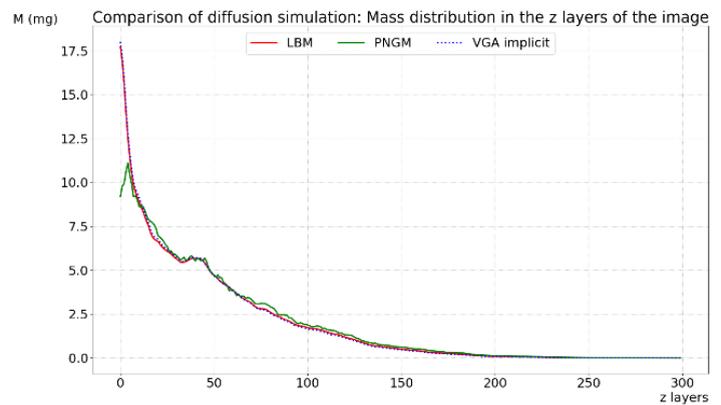

**FIGURE 7.** Comparison of diffusion simulations: implicit scheme of the GDE with a 30-second time step (blue line), implicit scheme of the PNGM with α=0.6 and a 15-second time step (green line), and LBM with a 0.43-second time step (red line).

Figure 7 illustrates a global comparison of the overall accuracy of each method. PNGM was simulated using the implicit scheme with a time step of 15 seconds, while GDE was solved using the implicit scheme with a time step of 30 seconds. In contrast, the time step for LBM is 0.43 seconds, calculated based on the 3D image resolution. The implicit scheme of GDE yields results similar to LBM and much more accurate results than the implicit scheme of PNGM.

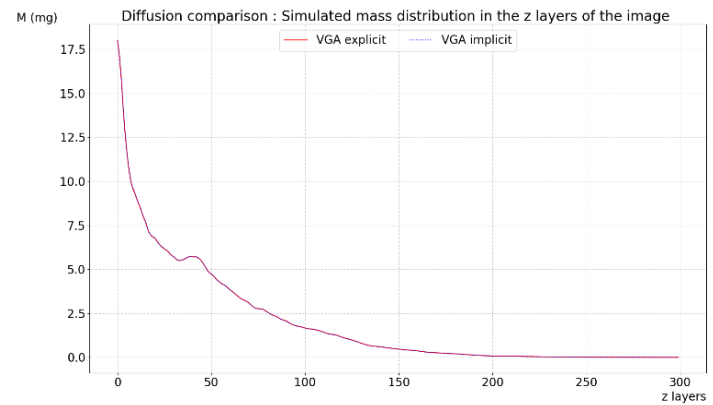

**FIGURE 8.** Comparison of diffusion simulation using the implicit and explicit schemes of the graph diffusion equation: the implicit scheme time step was set to 30 seconds, while the explicit scheme time step was set to 0.1 seconds.

The explicit scheme of the graph diffusion equation necessitates employing small time steps, resulting in computationally expensive calculations. Conversely, the implicit scheme allows for the use of larger time steps. Figure 8 illustrates a comparison of results between the implicit and explicit schemes, with the explicit scheme utilizing a time step of 0.1s and the implicit scheme employing a larger time step of 30s.

## C. Comparison of microbial decomposition simulations: VGA, LBM, and PNGM-based simulations

In this section, we validate the efficiency of our method, in simulating microbial decomposition of organic matter by comparison of simulation results to those obtained using LBM-based approach and comparing it to those obtained by PNGM-based method. We performed numerical simulations of microbial decomposition using the LBM-based approach, the VGA, and the ball network-based approach on the same dataset discussed previously.

The simulation procedure is as follows: Initially, we evenly distributed $M_2^0 = 289.5\ \mu g$ of DOM and introduced 1000 bacterial spots (clusters) into the pore space at random locations. This patchiness of bacterial distribution, as a result of cell growth mechanisms and environmental constraints, has been investigated in [35]. The clusters collectively represent $5.2 \times 10^7$ bacteria cells, which corresponds to $M_1^0 = 5.2 \times 10^7 \times 5.41 \times 10^{-8} \mu g = 2.8132\ \mu g$ of carbon.

For the PNGM-based approach, we distributed the mass $M_2^0$ homogeneously, (i.e., the same concentration in each ball), and we put bacterial mass into the balls corresponding to the random location of the spots.

The total mass in the pore space initially is as follows: $(M_1^0, M_2^0, M_3^0, M_4^0, M_5^0)$, where $M_3^0 = M_4^0 = M_5^0 = 0$, corresponding to 99.0376 % of dissolved organic matter and 0.962 % of living microorganisms.

The biological parameters employed in this study were adopted from [28] for Arthrobacter sp. 9R that was calibrated from real experimental data. These parameters are as follows:

- $\rho = 0.2\ day^{-1}$ is the relative respiration rate,
- $\mu = 0.5\ day^{-1}$ is the relative mortality rate,
- $\beta = 0.55$ is the proportion of microbial mass (MB) returning to DOM (the remaining portion is for MB returning to SOM),
- $v_{SOM} = 0.01\ day^{-1}$ and $v_{FOM} = 0.3\ day^{-1}$ are the relative decomposition rates for SOM and FOM respectively,
- $v_{DOM} = 9.6\ day^{-1}$ and $K_{DOM} = 0.001\ gC.g^{-1}$ are the maximum relative growth rate of MB, and the half saturation constant of DOM respectively.
- $D_{DOM} = 100950\ voxel^2.j^{-1}$ is the diffusion coefficient of DOM in water, with all other diffusion processes of other compounds canceled.

We conducted simulations using the three different methods over a 5-day period. For the LBM based method a time step of 0.43 seconds is used for diffusion, and the synchronous transformation processes with same time step. For the ball network model, we employed the calibrated implicit diffusion scheme from [20] and the asynchronous algorithm for transformation simulation with a 5-second time step.

For our approach, we conducted tests using different time steps to understand their impact on simulation results. Table 1 provides details on these time steps along with the elapsed time for each simulation. On a standard PC with an AMD Ryzen 7 PRO 6850H processor and 32.0 GB of RAM, the computation times are approximately as follows:

- LBM-based simulation: 3 weeks
- VGA simulation using an explicit scheme (test 3): 7.3 days
- VGA simulation using an implicit scheme (test 1): 5.2 days (refer to Table 1 for more details)
- Pore Network Model (PNGM)-based method: 45 minutes

Additionally, Table 1 presents details on the average intercorrelation between VGA simulations and LBM simulations.

**TABLE 1:** Tests of VGA for simulation of microbial decomposition of organic matter

| Test | | 1 | 2 | 3 | 4 |
|---|---|---|---|---|---|
| Figure | | 9 | 10 | 11 | 12 |
| Time step diffusion | Explicit | - | 0.1 (s) | 0.1 (s) | - |
| | Implicit | 1 (s) | - | - | 5 (s) |
| Time step transformation | | 1 (s) | 0.1 (s) | 0.43 (s) | 5 (s) |
| Computing time | | 5.3 (day) | 7.9 (day) | 6.4 (day) | 3.2 (day) |

The optimal intercorrelation was achieved in test 3, which is expected because it utilized the same time step as the LBM-based simulation, specifically 0.43 seconds for transformation. The decision to use a time step of 0.1 seconds for the explicit scheme was based on two factors: first, the accuracy of the explicit scheme, as it does not require approximation; second, 0.1 seconds represents the largest possible time step for the explicit scheme (at least for this dataset and experimental setup), otherwise we get negative values.

In contrast, Test 4 demonstrates that an implicit scheme with the same time step as that used for the pore network model (5s) produces results that are significantly divergent from both LBM and PNGM simulations (figure 12).

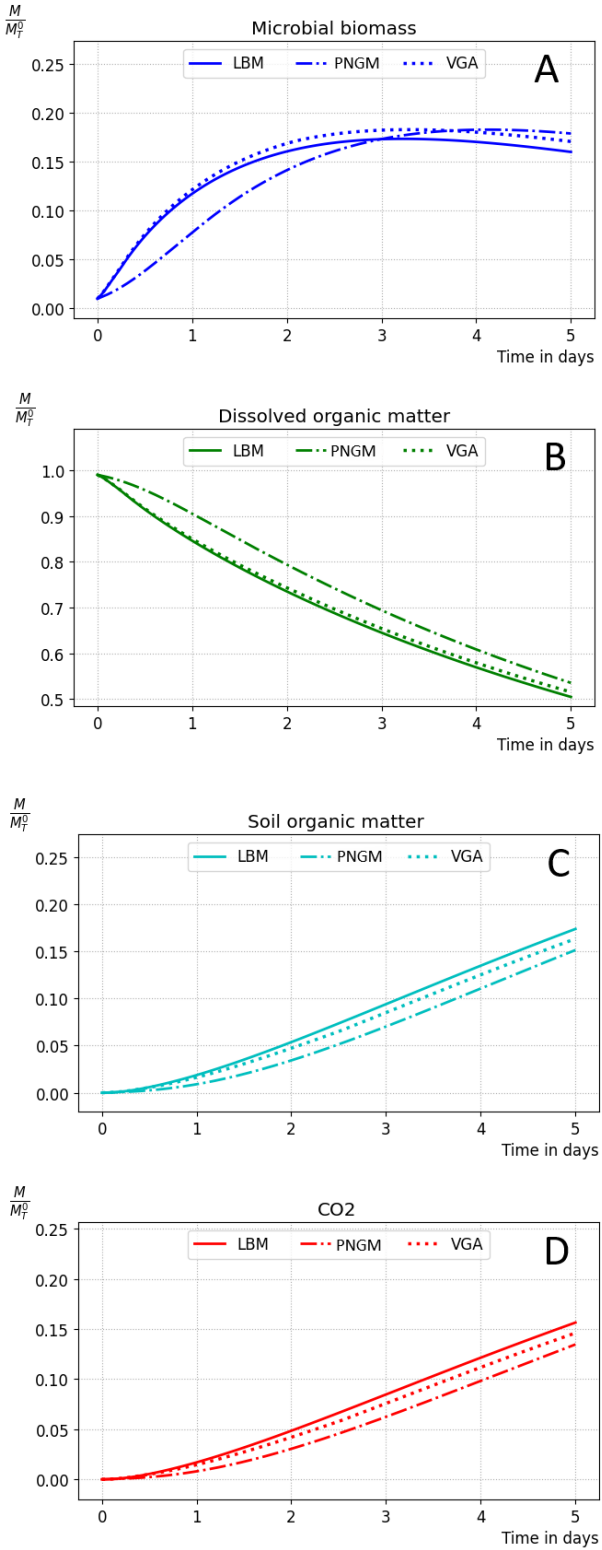

**FIGURE 9.** LBM-based approach using synchronous transformation with a time step of 0.43s, PNGM-based method using asynchronous transformation with 5s time step, VGA using implicit scheme and asynchronous transformation using same time step 1s.

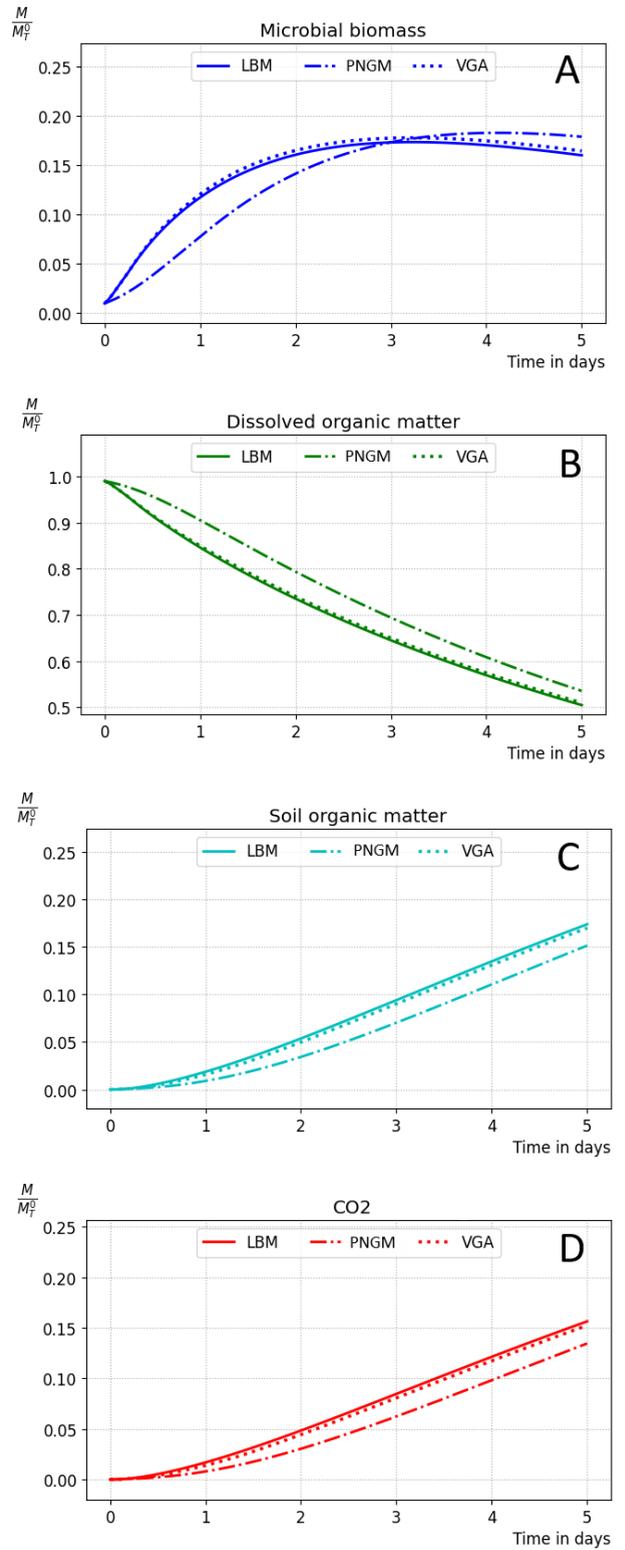

**FIGURE 10.** LBM-based approach using synchronous transformation with a time step of 0.43s, PNGM-based method using asynchronous transformation with 5s time step, VGA using explicit scheme and asynchronous transformation using same time step 0.1s.

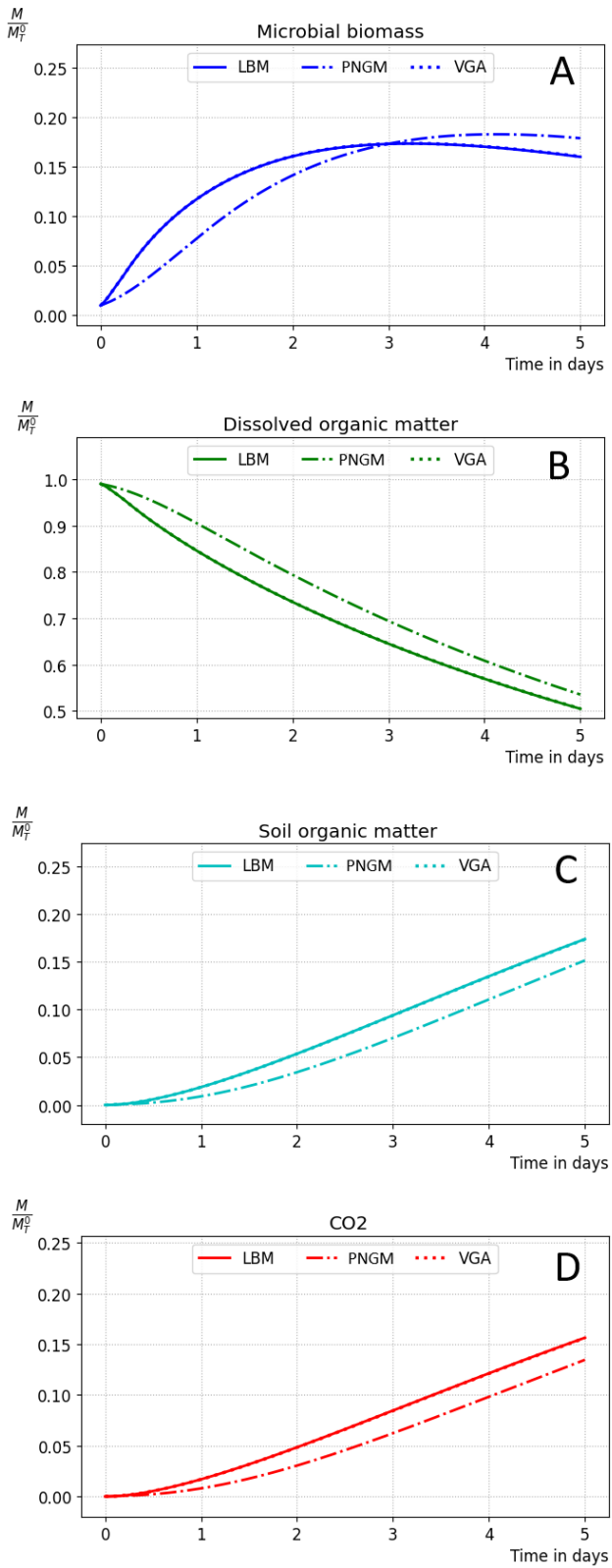

**FIGURE 11.** LBM-based approach using synchronous transformation with a time step of 0.43s, PNGM-based method using asynchronous transformation with a 5s time step, VGA using explicit scheme with a 0.1s time step and asynchronous transformation with a 0.43s time step.

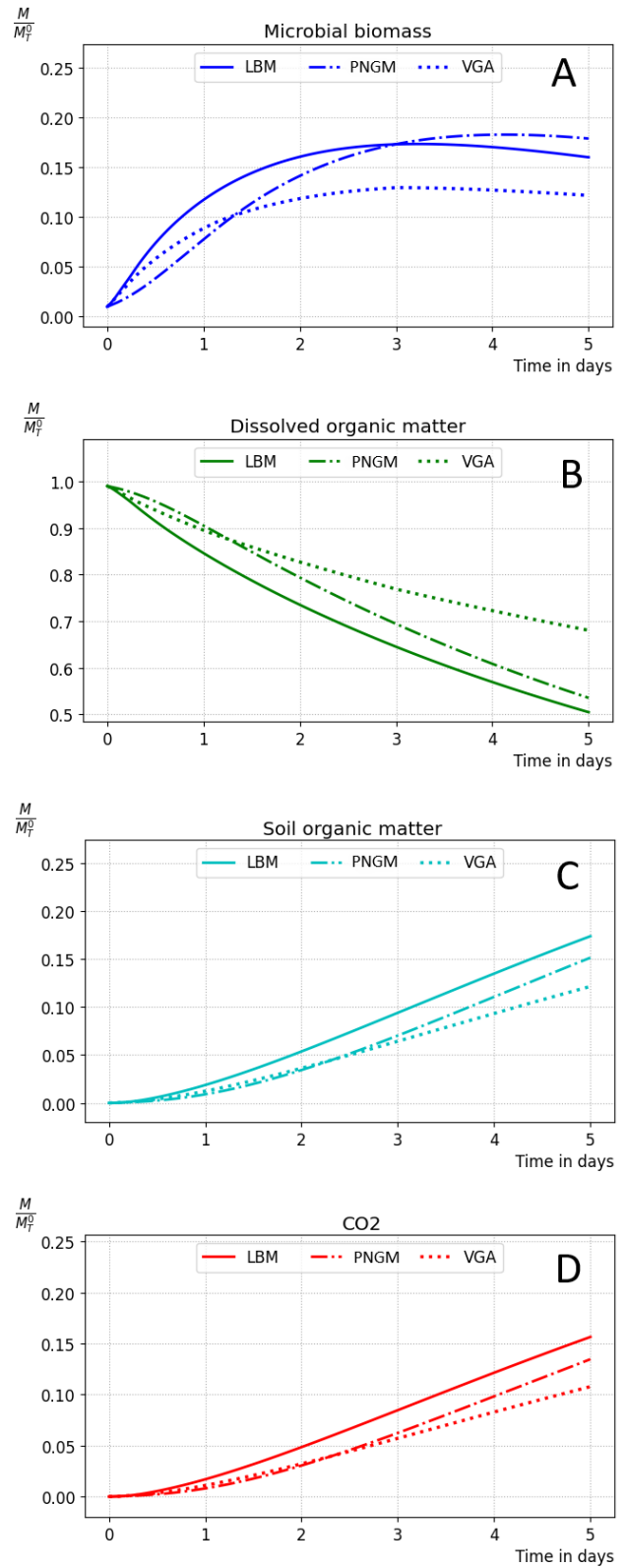

**FIGURE 12.** Microbial decomposition simulation: LBM-based approach using synchronous transformation with a time step of 0.43s, PNGM-based method using asynchronous transformation with a 5s time step, VGA using implicit scheme and asynchronous transformation using same time step of 5s.

The GDE-based method presents a practical alternative to LBM for simulation of microbial decomposition of organic matter in soil, as it yields comparable results at lower computational costs. The results obtained indicate that VGA outperforms the other two methods by striking a balance between accuracy and computational time. This superiority can be attributed to three main factors: firstly, the representation of the pore space using a graph of connected voxels, employing only a 6-connectivity policy, results in a much accurate representation than, the pore network models and a lighter representation compared to the lattices used in LBM (D3Q9 in this case). Secondly, simulating mass transport between connected voxels using Fick's law is computationally less expensive than simulating discrete collisions and particle propagation within a lattice grid. Finally, the implicit scheme of the graph diffusion equation, which allows the use of larger time steps than those used in LBM, is a key factor in reducing the number of iterations.

The PNGM-based approaches demonstrate significant cost-effectiveness compared to voxel-based approaches such as the LBM-based approach or the discussed voxel graph-based approach. Even time steps of 30 seconds still yield good results (refer to [20] for more details).

In the next section, a detailed theoretical approach will be provided for calculating the diffusional conductance coefficients for PNGM models using the voxel graph-based approach, in order to enhance accuracy.

## V. DIFFUSIONAL CONDUCTANCE COEFFICIENTS APPROXIMATION

### A. Principle and theoretical framework

In pore network geometrical modeling, pore space is approximated with a minimal set of maximal geometrical primitives (balls, ellipsoid, cylinders …). Then, a valuated graph is constructed from the set, where the adjacency is the geometrical adjacency between the geometrical primitives. For each pair of geometrically adjacent primitives, we need to determine the distance between gravitational centers and the surface of contact in order to simulate diffusion using Fick law. In previous works, we used a generalized coefficient that we multiply by the minimum of radii of the connected primitives, and we calibrate this generalized coefficient using LBM simulation [20].

In this subsection we discuss a possible way to determine approximatively the diffusional conductance coefficients of a pore network geometrical model.

Let $G_{VGA}(N_{VGA}, E_{VGA})$ be a voxel graph representation of a 3D image representing a pore space, where $N_{VGA} = \{v_1, \ldots, v_n\}$ represents the valid voxels of the image, $n$ is their number, and $E_{VGA} = \{(i,j) : i,j \in \{1, \ldots, n\} \text{ and } v_i \cap v_j \neq \emptyset\}$ encodes the adjacency between them.

Let $G_{PNGM}(N_{PNGM}, E_{PNGM})$ be the graph of connected geometrical primitives covering the pore space of the 3D image, where $N_{PNGM} = \{P_1, \ldots, P_q\}$ is the set of geometrical primitives, $q$ is their number, and $E_{PNGM} = \{(i,j) : i,j \in \{1, \ldots, n\} \text{ and } P_i \cap P_j \neq \emptyset\}$ encodes the adjacency between them. For all $k \in \{1, \ldots, q\}$ let $V(P_k) = \{v_i \in N_{VGA} : v_i \in P_k\}$ be the set of voxels contained within the primitive $P_k$. For theoretical explanation, suppose that:
$$k_1, k_2 \in \{1, \ldots, q\}, \quad V(P_{k_1}) \cap V(P_{k_2}) = \emptyset$$
and
$$\bigcup_{k \in \{1, \ldots, q\}} V(P_k) = N_{VGA}$$

Let $t \geq 0$ be a time, and $M_{PNGM}(t) = \{m^{PNGM}{}_1(t), \ldots, m^{PNGM}{}_q(t)\}$ be a mass distribution of the geometrical primitives $\{P_1, \ldots, P_q\}$. According to Fick's Law of diffusion, the flow of mass between two adjacent primitives $P_i$ and $P_j$, of volume $v_i$ and $v_j$ respectively, at time $t + \delta t$ is given by:

$$F_{i,j} = -Dc.\delta t.\alpha_{i,j}\left(\frac{m_i(t)}{v_i} - \frac{m_j(t)}{v_j}\right)$$

where $D_c$ is the diffusion coefficient, and $\alpha_{i,j}$ is what we call the diffusional conductance between the two primitives. This conductance typically depends on factors such as the contact surface between them, the distance between their centers of mass, the form of the geometrical primitives, and the difference in volume between the two primitives.

Then, the variation of mass distribution $\frac{dM_{PNGM}(t)}{dt}$ can be derived in the same way as we have done in Section 1.2, and it is given by:

$$\forall i \in \{1, \ldots, q\},$$

$$\frac{dm^{PNGM}{}_i(t)}{dt} = \sum_{j \in \vartheta(i)} -Dc.\alpha_{i,j}\left(\frac{m^{PNGM}{}_i(t)}{v_i} - \frac{m^{PNGM}{}_j(t)}{v_j}\right)$$

where $\vartheta(i) = \{j \in \{1, \ldots, q\} : P_i \cap P_j \neq \emptyset\}$ is the indexes of geometrically adjacent primitives to the primitive $P_i$. Numerically, solving the $PNGM\ diffusion\ model$, can be done by discretizing time using forward (explicit) or backward (implicit) Euler schemes.

Using the explicit scheme, we get:
$$\forall i \in \{1, \ldots, q\},$$

$$m^{PNGM}{}_i(t + \delta t) = m^{PNGM}{}_i(t) - Dc.\delta t. \sum_{j \in \vartheta(i)} .\alpha_{i,j}\left(\frac{m^{PNGM}{}_i(t)}{v_i} - \frac{m^{PNGM}{}_j(t)}{v_j}\right)$$

Then, we have

$$\forall i \in \{1, \ldots, q\}, \quad m^{PNGM^{k+1}}_i = m^{PNGM^k}_i - Dc.\delta t. \sum_{j \in \vartheta(i)} . \alpha_{i,j} \left( \frac{m^{PNGM^k}_i}{v_i} - \frac{m^{PNGM^k}_j}{v_j} \right) \quad (scheme\ 3)$$

Using the implicit scheme, we get:
$$\forall i \in \{1, \ldots, q\},$$
$$m^{PNGM}_i(t) = m^{PNGM}_i(t+\delta t) + Dc.\delta t. \sum_{j \in \vartheta(i)} . \alpha_{i,j} \left( \frac{m^{PNGM}_i(t+\delta t)}{v_i} - \frac{m^{PNGM}_j(t+\delta t)}{v_j} \right)$$

Then, we have

$$\forall i \in \{1, \ldots, q\}, m^{PNGM^k}_i = m^{PNGM^{k+1}}_i + Dc.\delta t. \sum_{j \in \vartheta(i)} . \alpha_{i,j} \left( \frac{m^{PNGM^{k+1}}_i}{v_i} - \frac{m^{PNGM^{k+1}}_j}{v_j} \right) \quad (scheme\ 4)$$

where $k$ is the iteration number.

The accuracy of the $PNGM\ diffusion\ model$ depends on the exactitude of calculating the coefficients $\{\alpha_{i,j} : (i,j) \in E_{PNM}\}$. In [2] and [7] the coefficients $\alpha_{i,j}$ were set to $\alpha \frac{S_{i,j}}{d_{i,j}}$ where $S_{i,j}$ is the contact surface between the primitive $P_i$ and the adjacent primitive $P_j$, $d_{i,j}$ is the distance between the centers of gravitation, and $\alpha$ is a unified coefficient for all adjacent primitives in the network that was calibrated using LBM to give the big intercorrelation between results of simulation.

In the following, we present a detailed approach to approximate the coefficients using a simple neural network and data generated using the voxel graph-based approach discussed before.

Let, $M_{VGA}(t) = \{m^{VGA}_1(t), \ldots, m^{VGA}_n(t)\}$ be the voxel distribution that correspond to the primitive's distribution $M_{PNM}(t)$ at time t, obtained by the voxel description $\{V(P_k): k \in \{1, \ldots, q\}\}$ of the primitives $\{P_k: k \in \{1, \ldots, q\}\}$.

The evolution of $M_{VGA}(t)$ over time can be calculated using the graph diffusion equation through the use of one of the schemes: the explicit (scheme 1) or the implicit (scheme 2). The framework for the simulation of $M_{VGA}(t)$ over time was discussed in section II and validated in section III.

Let $\{(X^p, Y^p): p \in \{1, \ldots, l\}\}$ be a dataset of couple of distributions: a random distribution $X^p$ of mass in the pore network and the corresponding next distribution $Y^p$ after a time step $\delta t$ of diffusion calculated using the voxel graph-based approach from the corresponding voxel description. Rewriting the scheme 3 for the dataset we have

$$\forall p \in \{1, \ldots, l\}, \forall i \in \{1, \ldots, q\}, y^p_i = x^p_i - Dc.\delta t. \sum_{j \in \vartheta(i)} . \alpha_{i,j} \left( \frac{x^p_i}{v_i} - \frac{x^p_i}{v_j} \right)$$

Finding $\{\alpha_{i,j} : (i,j) \in E_{PNGM}\}$ is equivalent to finding the set of parameters $\Theta = \{\theta_{i,j} : (i,j) \in E_{PNGM}\}$ that minimize for all data $\{(X^p, Y^p): p \in \{1, \ldots, l\}\}$ one of the following objectives:

$$L_1(\Theta) = \frac{1}{l} \sum_{p=1}^{l} \frac{1}{q} \sum_{i=1}^{q} \left( y^p_i - x^p_i + Dc.\delta t. \sum_{j \in \vartheta(i)} . \theta_{i,j} \left( \frac{x^p_i}{v_i} - \frac{x^p_i}{v_j} \right) \right)^2 \quad (Objective\ 1)$$

And

$$L_2(\Theta) = \frac{1}{l} \sum_{p=1}^{l} \frac{1}{q} \sum_{i=1}^{q} \left( x^p_i - y^p_i - Dc.\delta t. \sum_{j \in \vartheta(i)} . \theta_{i,j} \left( \frac{y^p_i}{v_i} - \frac{y^p_i}{v_j} \right) \right)^2 \quad (Objective\ 2)$$

The first objective calculates the L2 loss between the output of the explicit scheme 3 and the target (next time distribution) obtained using VGA.
The second objective minimizes the error between the input and second member of the implicit scheme applied to the target (next time distribution) obtained using VGA.
Stochastic Gradient Descent (SGD) is a fundamental optimization algorithm widely used to minimize such functions, particularly in the context of training machine learning models.

At its core, Gradient Descent is a technique for finding the minimum of a function by iteratively moving in the direction of the steepest decrease of the function. The key idea is to adjust the parameters of the model in small steps proportional to the negative of the gradient of the function with respect to those parameters.
 Standard Gradient Descent involves computing the gradient of the loss function with respect to all training examples, which can be computationally expensive, especially for large datasets. Stochastic Gradient Descent (SGD) offers a solution to this problem by updating the model's parameters more frequently using only a single data point or a small subset of data points at each step. This stochastic nature introduces noise into the optimization process, hence the name "Stochastic" Gradient Descent.

For an iteration of SGD, specifically for a set of data points $\{(X^p, Y^p): p \in \{1, ..., l\}\}$ (where p represents the size of the batch), the gradient for instance of the first objective is calculated using the following:

$$\frac{d}{d\theta_{i,j}} L_1(\Theta) = \frac{1}{l} \sum_{p=1}^{l} \frac{d}{d\theta_{i,j}} \left( \frac{1}{q} \sum_{m=1}^{q} \left( y^p_m - x^p_m + Dc.\delta t. \sum_{f \in \vartheta(m)} .\theta_{m,f} \left( \frac{x^p_m}{v_m} - \frac{x^p_f}{v_f} \right) \right)^2 \right)$$

Then,

$$\frac{d}{d\theta_{i,j}} L_1(\Theta) = \frac{1}{l} \sum_{p=1}^{l} \frac{1}{q} \sum_{m=1}^{q} \frac{d}{d\theta_{i,j}} \left( \left( y^p_m - x^p_m + Dc.\delta t. \sum_{f \in \vartheta(m)} .\theta_{m,f} \left( \frac{x^p_m}{v_m} - \frac{x^p_f}{v_f} \right) \right)^2 \right)$$

Then,

$$\frac{d}{d\theta_{i,j}} L_1(\Theta)$$

$$= \frac{1}{l} \sum_{p=1}^{l} \left[ \overbrace{\frac{1}{q} \sum_{\substack{m=1 \\ m \neq i}}^{q} \frac{d}{d\theta_{i,j}} \left( \left( y^p_m - x^p_m + Dc.\delta t. \sum_{f \in \vartheta(m)} .\theta_{m,f} \left( \frac{x^p_m}{v_m} - \frac{x^p_f}{v_f} \right) \right)^2 \right)}^{=0} \right.$$

$$\left. + \frac{1}{q} \frac{d}{d\theta_{i,j}} \left( \left( y^p_i - x^p_i + Dc.\delta t. \sum_{f \in \vartheta(i)} .\theta_{i,f} \left( \frac{x^p_i}{v_i} - \frac{x^p_f}{v_f} \right) \right)^2 \right) \right]$$

Then,

$$\frac{d}{d\theta_{i,j}} L_1(\Theta) = \frac{1}{l} \sum_{p=1}^{l} \left[ \frac{2}{q} \frac{d}{d\theta_{i,j}} \left( y^p_i - x^p_i + Dc.\delta t. \sum_{f \in \vartheta(i)} .\theta_{i,f} \left( \frac{x^p_i}{v_i} - \frac{x^p_f}{v_f} \right) \right) \times \left( y^p_i - x^p_i + Dc.\delta t. \sum_{f \in \vartheta(i)} .\theta_{i,f} \left( \frac{x^p_i}{v_i} - \frac{x^p_f}{v_f} \right) \right) \right]$$

Finally,

$$\frac{d}{d\theta_{i,j}} L_1(\Theta) = \frac{1}{l} \sum_{p=1}^{l} \left[ \frac{2}{q} Dc.\delta t \left( \frac{x^p_i}{v_i} - \frac{x^p_j}{v_j} \right) \times \left( y^p_i - x^p_i + Dc.\delta t \sum_{f \in \vartheta(i)} \theta_{i,f}. \left( \frac{x^p_i}{v_i} - \frac{x^p_f}{v_f} \right) \right) \right]$$

In a like manner the gradient of the second objective is calculated using the following:

$$\frac{d}{d\theta_{i,j}} L_2(\Theta) = \frac{1}{l} \sum_{p=1}^{l} \left[ \frac{2}{q} Dc.\delta t \left( \frac{y^p_i}{v_i} - \frac{y^p_j}{v_j} \right) \times \left( x^p_i - y^p_i - Dc.\delta t \sum_{f \in \vartheta(i)} \theta_{i,f}. \left( \frac{y^p_i}{v_i} - \frac{y^p_f}{v_f} \right) \right) \right]$$

In each training step, we calculate the set $\left\{ \frac{d}{d\theta_{i,j}} L(\Theta) : (i,j) \in E_{PNM} \right\}$, and we update the parameters using the following stochastic gradient descent policy:

$$\forall (i,j) \in E_{PNGM}, \quad \theta_{i,j} = \theta_{i,j} - lr \frac{\frac{d}{d\theta_{i,j}} L(\Theta)}{\left| \frac{d}{d\theta_{i,j}} L(\Theta) \right|}.$$

The use of the normalized gradient $\frac{\frac{d}{d\theta_{i,j}} L(\Theta)}{\left| \frac{d}{d\theta_{i,j}} L(\Theta) \right|}$ instead of the gradient $\frac{d}{d\theta_{i,j}} L(\Theta)$ addresses issues related to extremely small gradient magnitudes. This normalization helps in maintaining numerical stability, ensuring consistent and meaningful updates to the parameters, and potentially improving the convergence rate of the optimization process. Using the attained value of the objective function $L(\Theta)$ at each training iteration, we decide whether to stop training, continue, or adjust the hyperparameter "learning rate ($lr$)" and restart training.

After the training using stochastic gradient descent, we extract the learned parameters that will approximate the coefficient $\{\alpha_{i,j} : (i,j) \in E_{PNGM}\}$.

Note that the estimated diffusional conductance coefficients are theoretically applicable across all diffusion scenarios, regardless of the chosen diffusion coefficient or time step used in training, as they remain independent of these parameters.

Next subsection we test the discussed theoretical framework in predicting the diffusional conductance coefficient for the ball network model discussed before.

### B. Application example: Simulation of diffusion and microbial decomposition using PNGM

In the context of the ball network detailed in Section III, we initialize the parameters denoted as $\Theta = \{\theta_{i,j} : (i,j) \in E_{PNGM}\}$ by

$$\forall (i,j) \in E_{PNGM}, \quad \theta_{i,j} = \frac{S_{i,j}}{d_{i,j}}$$

Where $S_{i,j}$ represents the contact surface and $d_{i,j}$ signifies the distance between the two connected primitives $i$ and $j$. Subsequently, a synthetic 3D pore space image is generated to represent the balls, followed by the construction of the voxel graph as outlined in Section III. Thirty scenarios of Distribution of Organic Matter (DOM) within the pore space voxels are generated as follows: for each scenario, a random mass $M_0$ is generated and distributed randomly among the voxels. The diffusion process is simulated using the explicit scheme of the voxel graph diffusion equation with a time step of 0.1 seconds and a diffusion coefficient $D_c = 100950 \, voxel^2.j^{-1}$. The mass distribution in the voxels is recorded every 10 seconds during the simulation. Subsequently, all resulting voxel distributions are mapped to the corresponding ball distributions. These ball distributions are used to minimize the implicit objective $L_2$. A total of 3600 data points is obtained, each consisting of a mass distribution in the ball network and the corresponding mass distribution after 10 seconds of diffusion simulated using the accurate VGA-based simulation.

We run 1000 epochs stochastic gradient descent for training. Each epoch involves the selection of four random data points from the obtained dataset, then calculating normalized gradient and update parameters accordingly. Training starts with a learning rate of 0.1, which is then halved every 10 epochs.

During training, we calculate the $L_2$ loss applied to four chosen data points to track minimization. The history of the $L_2$ values is plotted in Figures 13.

The loss decreases significantly indicating rapid learning and convergence within the first 10 epochs.

Also, the training shows great stability and convergence to the solution, with a minimal $L_2$ value of $1.37766510 \times 10^{-8}$.

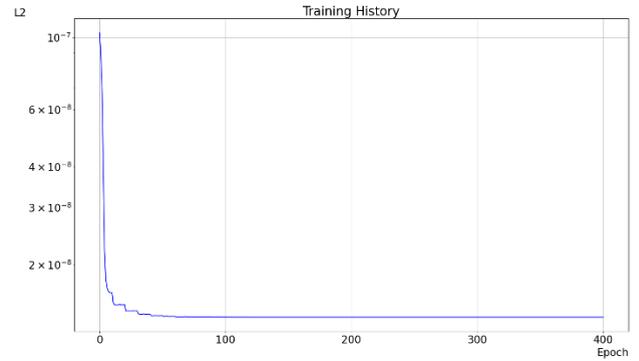

**FIGURE 13.** Training History: The x-axis represents the number of epochs, and the y-axis represents the L2 error for the four selected data points.

After epoch 355, the learning process stabilizes, and the stochastic gradient descent algorithm converges to the set $\Theta^* = \{\theta_{i,j}^* : (i,j) \in E_{PNGM}\}$ that minimize the application $\Theta \mapsto L_2(\Theta)$. The obtained coefficients replace the coefficients $\{\alpha_{i,j} : (i,j) \in E_{PNGM}\}$ in one of the schemes (either Scheme 3 or Scheme 4) to simulate diffusion within the ball network, which serves as an approximation of the pore space.

Since the number of balls is significantly smaller than the number of voxels constituting the pore space, simulations using balls are computationally more efficient compared to those using voxels. Accurate determination of diffusional conductance coefficients for the ball network is crucial for precise diffusion simulation. Figures 14 and 15 illustrate a comparison of simulation results using two different sets of diffusional coefficients: the old coefficients from prior studies and the new coefficients derived using the SGD and data generated by the voxel approach. These results are compared against the accurate VGA-based simulations.

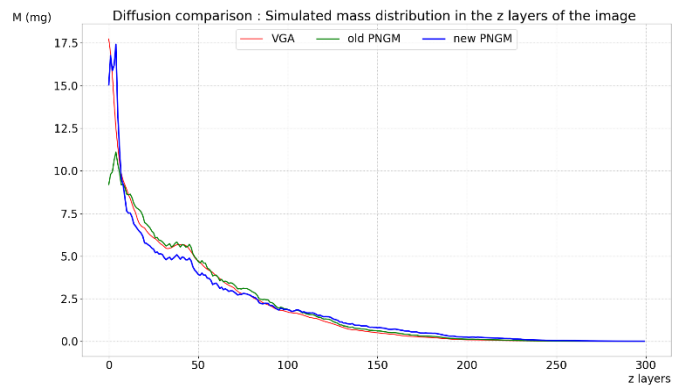

**FIGURE 14.** Diffusion simulation comparison: VGA-based simulation (red curve), PNGM-based simulation calibrated by LBM-based simulation (green curve), and PNGM-based simulation with the approximated diffusional conductance coefficients (blue curve).

Figure 14 presents a comparison of simulated mass after 1.73 hours of diffusion, using the same scenario described in Section III, with both VGA-based and PNGM-based simulations. The figure includes results from VGA using explicit scheme with a 0.1s time step a PNGM model calibrated by LBM simulation (referred to as the old PNGM) and from a PNGM model with diffusional conductance coefficients obtained after 1000 epochs of training (referred to as the new PNGM) using explicit scheme with a 0.1s time step. The new PNGM shows an improvement over the old PNGM, demonstrating enhanced accuracy in the simulation results.

We conducted microbial decomposition simulations under the same scenario as in Section III, employing the old PNGM-based simulation calibrated using LBM in previous work [20], and the improved PNGM-based simulation using the newly obtained diffusional conductance coefficients. These simulations were done using the implicit scheme with a 10s-time step and the asynchronous transformation using a 30s-time step. The results of these simulations are compared with those of the VGA-based simulation using explicit scheme with a 0.1s-time step, and asynchronous transformation with a 0.43s-time step.

The improved accuracy of the new PNGM model is evident from its closer alignment with the VGA-based simulation results. Figure 15 compares simulated microbial decomposition using the same scenario outlined in Section III, with both LBM-based and PNGM-based simulations. The figure includes results from a PNGM model calibrated by LBM simulation (old PNGM) and from a PNGM model with diffusional conductance coefficients obtained after training (new PNGM).

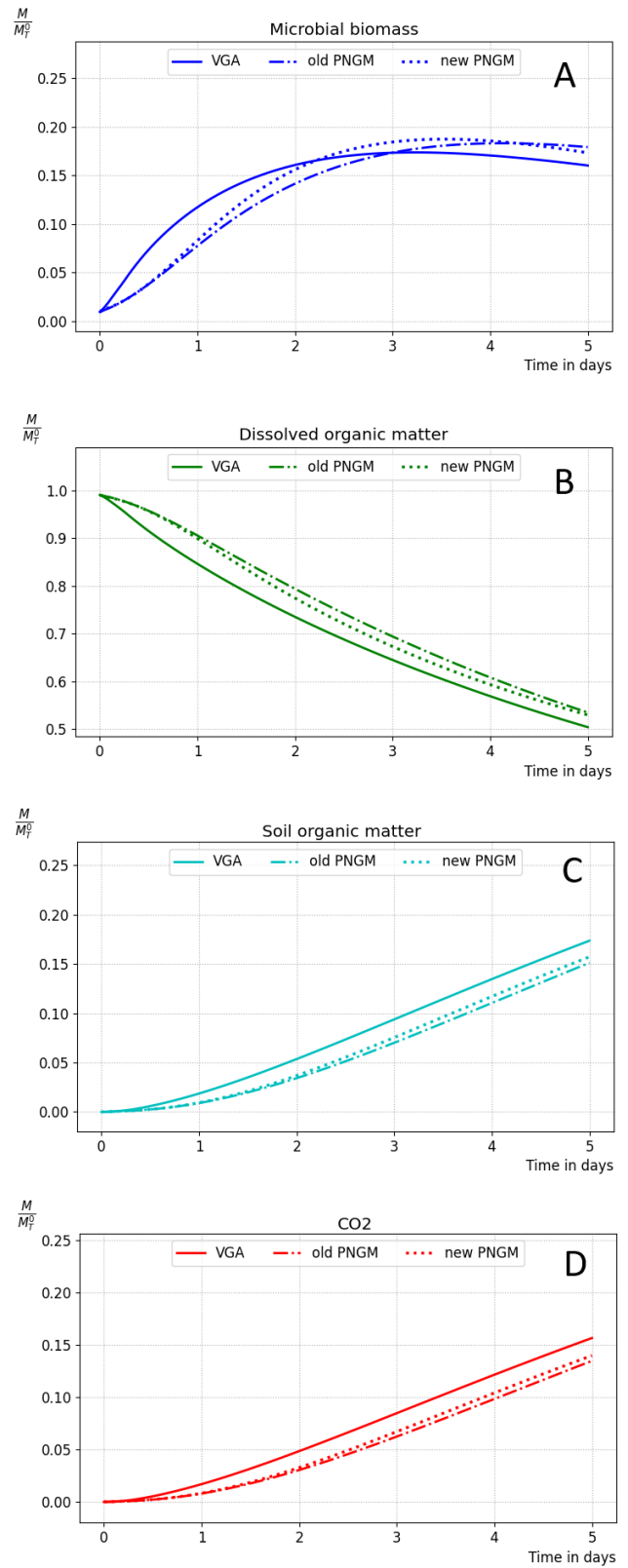

**FIGURE 15.** Microbial decomposition simulation: VGA-based simulation (solid lines), PNGM-based simulation calibrated by LBM-based simulation (dash-dotted lines), and PNGM-based simulation with the approximated diffusional conductance coefficients (dotted lines).

## VI. Perspectives and advantages
The following outlines the primary opportunities facilitated by employing the voxel graph-based framework:
- <u>Refining diffusional conductance coefficients:</u> due to its accuracy and minimal computational expense, the method is employed to generate training data for calibrating diffusion between connected primitives in PNGM-based simulations. Thereby increasing precision and accuracy.
- <u>Incorporating pore space deformation:</u> the PNGM-based approaches are very limited in modeling deformation of the pore space as the geometric primitives of the PNGM cannot be deformed or intersect. In the present paper, we do not consider any deformation of the pore space. Nevertheless, through the representation of pore space as a graph of connected voxels, our simulations can now include the deformations of pore structures during modeling by considering the movement or the elimination of voxels.
- <u>Upscaling:</u> Using graph neural networks, coupled with the voxel graph representation of the pore space, we can train neural models to mimic the discussed model or to learn intrinsic patterns in the microbial dynamics in order to predict long-term dynamics of microbial behavior from 3D images of larger samples that will be a key point to address upscaling issues.
- The method can be used to simulate a wide range of transformation diffusion processes in porous media or any fractured environment.

## VII. Conclusion
In conclusion, this paper introduces a Voxel Graph-based Approach (VGA) for simulating microbial activity in soil pores, addressing limitations encountered by traditional methods such as huge computational cost in LBM and accuracy in PNGMs.

By representing soil pore spaces as a graph of connected voxels and modeling transport using Fick's law and parallel implementations, VGA offers an enhanced accuracy and computational efficiency in simulating mass transport and microbial decomposition processes and can be used to model and simulate a wide range of reaction-transport dynamics in complex shapes.

Our comparative analysis with LBM and PNGM demonstrates that VGA achieves similar results to LBM and outperform PNGM in terms of accuracy.

While PNGM is generally less time-consuming than both methods, we showed that the great balance between accuracy and computation complexity in VGA makes the approximation of diffusional conductance coefficients between connected primitives in PNGM possible through stochastic gradient descent, hence enhancing its accuracy. This improvement can play a significant role in predicting the long-term behavior of the microbial system in larger samples in a reasonable time, which can address upscaling issues.

Although the high gain in computing time using PNGM, including deformation of soil pores, is impossible due to geometrical primitives intersecting or deforming, VGA, through modeling the pore space using connected voxels, can include deformation processes easily by moving or removing voxels.

Future work will focus on modeling the deformation of the pore space and simulating the coupled deformation and reaction-transport dynamics in the complex geometry of soil.

Necessarily, integrating more complex biological interactions and expanding applications to real-world environmental scenarios will surely contribute to sustainable soil management and ecosystem preservation efforts.

## ACKNOWLEDGMENT
The present research was made possible through the French Institute of Research for Development's international doctoral program scholarship offered to PhD student Mouad KLAI and through the I-Maroc project (APRD program) and the Microlarge project (French ANR).

## APPENDIX A: Synchronous and Asynchronous schemes
Algorithm 1: Asynchronous transformation
Inputs:
$(X_i)_{i \in N}$ biological attributes of the nodes,
$\delta t$ time step,
$\rho, \mu, \beta, v_{SOM}, v_{FOM}, v_{DOM}, K_{DOM}$ the biological parameters of the model,
Procedure:
start;
Parallel loop: for $i$ in N:
$uptake \leftarrow \frac{v_{DOM}.x_i^2}{K_{DOM}+x_i^2}.x_i^1.\delta t$ ; // Calculate bacterial uptake of DOM
if $uptake > x_i^2$:
$uptake \leftarrow x_i^2$ ; // Adjust uptake if demand exceeds available DOM
end if;
$resp \leftarrow \rho.x_i^1.\delta t$ ; // Calculate bacterial respiration
$morta \leftarrow \mu.x_i^1.\delta t$ ; // Calculate bacterial mortality
if $resp + morta > x_i^1$:
if $morta > x_i^1$:
$morta \leftarrow x_i^1$ ;
$resp \leftarrow 0$ ;
else:
$resp \leftarrow x_i^1 - morta$ ;
end if;
end if;

$mortaSOM \leftarrow (1-\beta).morta$ ; // Calculate mortality contributing to SOM
$mortaDOM \leftarrow \beta.morta$ ; // Calculate mortality contributing to DOM
$turnFOM \leftarrow v_{FOM}.x_i^4.\delta t$ ; // Calculate FOM turnover
if $turnFOM > x_i^4$:
$turnFOM \leftarrow x_i^4$;
end if;
$turnSOM \leftarrow v_{SOM}.x_i^3.\delta t$ ;// Calculate SOM turnover
If $turnSOM > x_i^3$:
$turnSOM \leftarrow x_i^3$;
end if;
$x_i^1 \leftarrow x_i^1 + (uptake - morta - resp)$ ; // Update active bacterial mass
$x_i^2 \leftarrow x_i^2 + (morDOM - uptake + turnFOM + turnSOM)$ ; // Update DOM mass
$x_i^3 \leftarrow x_i^3 + (morSOM - turnSOM)$; // Update SOM mass
$x_i^4 \leftarrow x_i^4 - turnFOM$ ;// Update FOM mass
$x_i^5 \leftarrow x_i^5 + resp$; // Update respiration product
end for;
end;

Algorithm 2: Synchronous transformation
Inputs:
$(X_i)_{i \in N}$ biological attributes of the nodes,
$\delta t$ time step,
$\rho, \mu, \beta, v_{SOM}, v_{FOM}, v_{DOM}, K_{DOM}$ the biological parameters of the model,
Procedure:
start;
Parallel loop: for $i$ in N:
if $x_i^1 > 0$:
if $x_i^2 > 0$:
$uptake \leftarrow \frac{v_{DOM}.x_i^2}{K_{DOM}+x_i^2}.x_i^1.\delta t$ ; // Calculate bacterial uptake of DOM
if $uptake < x_i^2$:
$x_i^1 \leftarrow x_i^1 + uptake$ ; // let microorganisms grow
$x_i^2 \leftarrow x_i^2 - uptake$ ; // consume DOM
else:
$x_i^1 \leftarrow x_i^1 + x_i^2$ ;// let microorganisms consume all available DOM
$x_i^2 \leftarrow 0$; // all DOM is consumed
end if;
end if;
$morta \leftarrow \mu.x_i^1.\delta t$ ; // Calculate bacterial mortality
if $morta < x_i^1$:
$x_i^1 \leftarrow x_i^1 - morta$; //microorganisms' mortality
$x_i^2 \leftarrow x_i^2 + \beta.morta$ ; // fast decomposition
$x_i^3 \leftarrow x_i^3 + (1-\beta).morta$ ; // slow decomposition
else:
$x_i^2 \leftarrow x_i^2 + \beta.x_i^1$; // fast decomposition
$x_i^3 \leftarrow x_i^3 + (1-\beta).x_i^1$ ; // slow decomposition
end if;
$resp \leftarrow \rho.x_i^1.\delta t$ ; // Calculate bacterial mortality

if $resp < x_i^1$:
$x_i^1 \leftarrow x_i^1 - resp$; //microorganisms' respiration
$x_i^5 \leftarrow x_i^5 + resp$ ; // $CO_2$ production
else:
$x_i^5 \leftarrow x_i^5 + x_i^1$; // $CO_2$ production
$x_i^1 \leftarrow 0$; //microorganisms' respiration
end if;
end if;
if $x_i^3 > 0$:
$turnSOM \leftarrow v_{SOM}.x_i^3.\delta t$ ;// Calculate SOM turnover
if $resp < x_i^3$:
$x_i^3 \leftarrow x_i^3 - turnSOM$; //SOM turnover
$x_i^2 \leftarrow x_i^2 + turnSOM$; // SOM to DOM
else:
$x_i^3 \leftarrow 0$;
$x_i^2 \leftarrow x_i^2 + x_i^3$;
end if;
end if;
if $x_i^4 > 0$:
$turnFOM \leftarrow v_{FOM}.x_i^4.\delta t$ ;// Calculate FOM turnover
if $resp < x_i^3$:
$x_i^4 \leftarrow x_i^4 - turnFOM$; //FOM turnover
$x_i^2 \leftarrow x_i^2 + turnFOM$; // FOM to DOM
else:
$x_i^4 \leftarrow 0$;
$x_i^2 \leftarrow x_i^2 + x_i^4$;
end if;
end if;
end for;
end;

**REFERENCES**

[1] L. Philippot, J. M. Raaijmakers, P. Lemanceau, and W. H. Van Der Putten, "Going back to the roots: the microbial ecology of the rhizosphere," Nature Reviews Microbiology, vol. 11, no. 11, pp. 789-799, Nov. 2013.
[2] R. D. Bardgett and W. H. Van Der Putten, "Belowground biodiversity and ecosystem functioning," Nature, vol. 515, no. 7528, pp. 505-511, Nov. 2014.
[3] N. Fierer, "Embracing the unknown: disentangling the complexities of the soil microbiome," Nature Reviews Microbiology, vol. 15, no. 10, pp. 579-590, Oct. 2017.
[4] A. Juyal, T. Eickhorst, R. Falconer, P. C. Baveye, A. Spiers, and W. Otten, "Control of pore geometry in soil microcosms and its effect on the growth and spread of Pseudomonas and Bacillus sp.," Frontiers in Environmental Science, vol. 6, p. 73, Jul. 2018.
[5] J. A. Dungait, D. W. Hopkins, A. S. Gregory, and A. P. Whitmore, "Soil organic matter turnover is governed by accessibility not recalcitrance," Global Change Biology, vol. 18, no. 6, pp. 1781-1796, Jun. 2012.
[6] A. N. Kravchenko and G. P. Robertson, "Whole-profile soil carbon stocks: The danger of assuming too much from analyses of too little," Soil Science Society of America Journal, vol. 75, no. 1, pp. 235-240, Jan. 2011.
[7] B. Mbé, O. Monga, V. Pot, W. Otten, F. Hecht, X. Raynaud, N. Nunan, C. Chenu, P. C. Baveye, and P. Garnier, "Scenario modelling of carbon mineralization in 3D soil architecture at the microscale: Toward an accessibility coefficient of organic matter for bacteria," European Journal of Soil Science, vol. 73, no. 1, pp. e13144, Jan. 2022.
[8] X. Portell, V. Pot, P. Garnier, W. Otten, and P. C. Baveye, "Microscale heterogeneity of the spatial distribution of organic matter can promote

bacterial biodiversity in soils: insights from computer simulations," Front. Microbiol., vol. 9, p. 368481, Jul. 2018.

[9] P. C. Baveye, L. S. Schnee, P. Boivin, M. Laba, and R. Radulovich, "Soil organic matter research and climate change: merely re-storing carbon versus restoring soil functions," *Frontiers in Environmental Science*, vol. 8, Art. no. 579904, Sep. 10, 2020.

[10] P. C. Baveye, W. Otten, A. Kravchenko, M. Balseiro-Romero, É. Beckers, M. Chalhoub, C. Darnault, T. Eickhorst, P. Garnier, S. Hapca, and S. Kiranyaz, "Emergent properties of microbial activity in heterogeneous soil microenvironments: different research approaches are slowly converging, yet major challenges remain," *Frontiers in Microbiology*, vol. 9, Art. no. 367238, Aug. 27, 2018.

[11] H. Sulieman, M.S. Jouini, M. Alsuwaidi, E.W. Al-Shalabi, O.A. Al Jallad, "Multiscale investigation of pore structure heterogeneity in carbonate rocks using digital imaging and SCAL measurements: A case study from Upper Jurassic limestones, Abu Dhabi, UAE" PLoS ONE, vol. 19, no. 2, pp. e0295192, 2024.

[12] S. Matsumura, A. Kondo, K. Nakamura, T. Mizutani, E. Kohama, K. Wada, T. Kobayashi, N. Roy, and J. D. Frost, "3D image scanning of gravel soil using in-situ X-ray computed tomography," Scientific Reports, vol. 13, no. 1, p. 20007, Nov. 2023.

[13] P. V., X. Portell, W. Otten, P. Garnier, O. Monga, and P. C. Baveye, "Understanding the joint impacts of soil architecture and microbial dynamics on soil functions: Insights derived from microscale models," Eur. J. Soil Sci., vol. 73, no. 3, pp. e13256, May 2022.

[14] A. M. Tartakovsky, P. Meakin, T. D. Scheibe, and B. D. Wood, "A smoothed particle hydrodynamics model for reactive transport and mineral precipitation in porous and fractured porous media," Water Resources Research, vol. 43, no. 5, May 2007.

[15] H. Yoon, A. J. Valocchi, C. J. Werth, and T. Dewers, "Pore-scale simulation of mixing-induced calcium carbonate precipitation and dissolution in a microfluidic pore network," Water Resources Research, vol. 48, no. 2, Feb. 2012.

[16] L. Li, C. A. Peters, and M. A. Celia, "Upscaling geochemical reaction rates using pore-scale network modeling," Advances in Water Resources, vol. 29, no. 9, pp. 1351-1370, Sep. 2006.

[17] A. Genty and V. Pot, "Numerical simulation of 3D liquid–gas distribution in porous media by a two-phase TRT lattice Boltzmann method," Transp. Porous Media, vol. 96, pp. 271-294, Jan. 2013.

[18] V. Pot, S. O. Peth, O. Monga, L. E. Vogel, A. Genty, P. Garnier, L. Vieublé-Gonod, M. Ogurreck, F. Beckmann, and P. C. Baveye, "Three-dimensional distribution of water and air in soil pores: comparison of two-phase two-relaxation-times lattice-Boltzmann and morphological model outputs with synchrotron X-ray computed tomography data," *Advances in Water Resources*, vol. 84, pp. 87-102, Oct. 1, 2015.

[19] L. E. Vogel, D. Makowski, P. Garnier, L. Vieublé-Gonod, Y. Coquet, X. Raynaud, N. Nunan, C. Chenu, R. Falconer, and V. Pot, "Modeling the effect of soil meso-and macropores topology on the biodegradation of a soluble carbon substrate," Adv. Water Resour., vol. 83, pp. 123-136, Sep. 2015.

[20] O. Monga, F. Hecht, M. Serge, M. Klai, M. Bruno, J. Dias, P. Garnier, and V. Pot, "Generic tool for numerical simulation of transformation-diffusion processes in complex volume geometric shapes: Application to microbial decomposition of organic matter," Comput. Geosci., vol. 169, p. 105240, Dec. 2022.

[21] O. Monga, F. N. Ngom, and J. F. Delerue, "Representing geometric structures in 3D tomography soil images: Application to pore-space modeling," Comput. Geosci., vol. 33, no. 9, pp. 1140-1161, Sep. 2007.

[22] D. Silin and T. Patzek, "Pore space morphology analysis using maximal inscribed spheres," Physica A: Stat. Mech. Appl., vol. 371, no. 2, pp. 336-360, Nov. 2006.

[23] H. J. Vogel, J. Tölke, V. P. Schulz, M. Krafczyk, and K. Roth, "Comparison of a lattice-Boltzmann model, a full-morphology model, and a pore network model for determining capillary pressure–saturation relationships," Vadose Zone J., vol. 4, no. 2, pp. 380-388, May 2005.

[24] N. Misaghian, M. Agnaou, M. A. Sadeghi, H. Fathiannasab, I. Hadji, E. Roberts, and J. Gostick, "Prediction of diffusional conductance in extracted pore network models using convolutional neural networks," Comput. Geosci., vol. 162, p. 105086, May 2022.

[25] T. M. Mayhew, "Estimating oxygen diffusive conductances of gas-exchange systems: A stereological approach illustrated with the human placenta," Ann. Anat.-Anat. Anz., vol. 196, no. 1, pp. 34-40, Jan. 2014.

[26] T. B. Boving and P. Grathwohl, "Tracer diffusion coefficients in sedimentary rocks: correlation to porosity and hydraulic conductivity," J. Contam. Hydrol., vol. 53, no. 1-2, pp. 85-100, Dec. 2001.

[27] A. Fick, "On liquid diffusion," J. Membr. Sci., vol. 100, no. 1, pp. 33-38, Mar. 1995.

[28] O. Monga, P. Garnier, V. Pot, E. Coucheney, N. Nunan, W. Otten, and C. Chenu, "Simulating microbial degradation of organic matter in a simple porous system using the 3-D diffusion-based model MOSAIC," Biogeosciences, vol. 11, no. 8, pp. 2201-2209, Apr. 2014.

[29] A. N. Houston, S. Schmidt, A. M. Tarquis, W. Otten, P. C. Baveye, and S. M. Hapca, "Effect of scanning and image reconstruction settings in X-ray computed microtomography on quality and segmentation of 3D soil images," Geoderma, vol. 207, pp. 154-165, Oct. 2013.

[30] N. Nunan, K. Ritz, D. Crabb, K. Harris, K. Wu, J. W. Crawford, and I. M. Young, "Quantification of the in situ distribution of soil bacteria by large-scale imaging of thin sections of undisturbed soil," FEMS Microbiol. Ecol., vol. 37, no. 1, pp. 67-77, Aug. 2001.